\newcommand{\CLASSINPUTinnersidemargin}{18mm}
\newcommand{\CLASSINPUToutersidemargin}{12mm}
\newcommand{\CLASSINPUTtoptextmargin}{20mm}
\newcommand{\CLASSINPUTbottomtextmargin}{25mm}
\documentclass[10pt,conference,a4paper]{IEEEtran}

\usepackage[utf8]{inputenc}

\usepackage{times}

\usepackage{graphicx}
\usepackage{subcaption}

\DeclareGraphicsExtensions{.png,.eps,.ps,.pdf}

\usepackage{xcolor,soul,framed} 

\colorlet{shadecolor}{yellow}

\usepackage{array}
\usepackage{mdwmath}
\usepackage{mdwtab}
\usepackage{eqparbox}
\usepackage{url}
\usepackage{cite}
\usepackage{amsmath,amssymb,amsfonts}
\usepackage{algorithmic}
\usepackage[ruled,vlined]{algorithm2e}
\usepackage{multirow}
\usepackage{textcomp}

\begin{document}

\title{Temporal graph-based approach for behavioural entity classification}

\author{\IEEEauthorblockN{Francesco Zola}
\IEEEauthorblockA{Vicomtech Foundation,\\Basque Research and\\Technology Alliance (BRTA)\\20009 San Sebastian, Spain\\fzola@vicomtech.org}
\and
\IEEEauthorblockN{Lander Segurola}
\IEEEauthorblockA{Vicomtech Foundation,\\Basque Research and\\Technology Alliance (BRTA)\\20009 San Sebastian, Spain\\lsegurola@vicomtech.org}
\and
\IEEEauthorblockN{Jan Lukas Bruse}
\IEEEauthorblockA{Vicomtech Foundation,\\Basque Research and\\Technology Alliance (BRTA)\\ 20009 San Sebastian, Spain\\jbruse@vicomtech.org}
\and
\IEEEauthorblockN{Mikel Galar Idoate}
\IEEEauthorblockA{Institute of Smart Cities,\\Public University of\\ Navarre,\\ 31006 Pamplona, Spain\\mikel.galar@unavarra.es}}

\maketitle

\begin{abstract}
Graph-based analyses have gained a lot of relevance in the past years due to their high potential in describing complex systems by detailing the actors involved, their relations and their behaviours. Nevertheless, in scenarios where these aspects are evolving over time, it is not easy to extract valuable information or to characterize correctly all the actors.

In this study, a two phased approach for exploiting the potential of graph structures in the cybersecurity domain is presented. The main idea is to convert a network classification problem into a graph-based behavioural one. We extract these graph structures that can represent the evolution of both normal and attack entities and apply a temporal dissection approach in order to highlight their micro-dynamics. Further, three clustering techniques are applied to the normal entities in order to aggregate similar behaviours, mitigate the imbalance problem and reduce noisy data. Our approach suggests the implementation of two promising deep learning paradigms for entity classification based on Graph Convolutional Networks.

\end{abstract}


{\bf Tipo de contribución:}  {\it Investigación en desarollo}

\section{Introduction}
Cybersecurity has become a critical aspect for many companies as a vulnerability or a breach in its infrastructure can generate a considerable loss of value, in economical, reputational, digital, psychological and societal terms\cite{agrafiotis2018taxonomy}. The increasing amount of cyber threats and their significant impacts, has led to increased efforts for preventing and reducing their risks \cite{khraisat2019survey}. In this sense, machine learning and derived techniques such as deep learning are promising means to obtain new insights from available data in order to quantify cyber risks and to optimize cybersecurity operations \cite{sarker2020cybersecurity}. 

These techniques are principally used for performing anomaly detection \cite{omar2013machine}, an operation that can be used for risk detection as well as a preventive approach. In particular, anomaly detection has shown very promising results analyzing data as graphs \cite{akoglu2015graph}. Graph structures help integrate both structured and unstructured data in a representation of entities and the relationships among them.

Nevertheless, in real use cases, entity behaviour can evolve such as entities disappearing, emerging or simply changing their dynamics. In these cases, considering a static graph is not sufficient to describe a complex system, not least as with this monolithic approach small interactions between entities can be obscured by more frequent ones - generating a loss of classification quality. In such a skewed scenario, machine learning models tend to focus on macro-dynamics and may falsely assign one static behaviour for each entity as generated by the overall status.

In this work, we propose a two phased approach for addressing these issues. In the first phase, called \textit{Manipulation phase}, graph-based structures are extracted from the initial network traffic dataset defining key concepts such as entity, edges and entity behaviour using a temporal dissection approach. Then, still in the first phase, a clustering operation is applied for reducing noisy data and addressing the class imbalance problem. In the second phase, called \textit{Learning phase}, deep learning models are trained with these graph data in order to detect attack/malicious behavioural entities.
In this work, operations related to the \textit{Manipulation phase} are introduced and three clustering techniques are compared using the UNSW15 dataset \cite{moustafa2015unsw}. Further, detailed guidelines for the \textit{Learning phase} are provided.

\section{Methodology}\label{sec:proposal}

\subsection{Problem description}\label{sec:problem}
In several cases, when the aim is to perform a network traffic classification, the most common and straightforward idea is to consider the information as a time series and then apply Artificial Intelligence paradigms for classifying these flows in order to discover malicious/anomalous patterns.
In this study, a two phased approach based on converting this time series analysis into a graph-based behavioural classification is proposed.

During the first phase, \textit{Manipulation phase}, the initial dataset is manipulated in order to extract graph information. The main idea is to split this initial dataset in fixed time interval called "snapshot", and, in each of them, extract graph-based structures defining entities and edges. In this sense, it is important to perform this temporal dissection operation in order to uncover micro-dynamics and highlight how the entities relations change within the time. 
Once the graph structures are ready, a clustering algorithm is applied in order to aggregate similar behaviours, i.e. similar entities, for reducing the overall amount of information. This operation is very important, especially when the initial dataset is characterized by noisy data or it is affected by a strong imbalanced problem. In these cases, applying the clustering only over the majority class helps to reduce those problems. In this research, $3$ clustering techniques are compared: Density-Based Spatial Clustering of Applications with Noise (DBSCAN \cite{ester1996density}), Ordering Points To Identify the Cluster Structure (OPTICS \cite{ankerst1999optics}) and Hierarchical DBSCAN (HDBSCAN\cite{campello2013density})
All these operations allow to create the appropriate input for the next phase (\textit{Learning phase}), which is focused on training $2$ graph deep learning models, both belonged to the Graph Convolutional Network (GCN) family, for classifying the node behaviours and detect the malicious/anomalous ones.

It is important to remark that this work is not focused on classifying or predicting directly the node's communication as presented in \cite{oba2020graph, zheng2019addgraph}, neither on developing another framework based on graph deep learning models \cite{jiang2019anomaly}, but the key idea is to convert network flow classification into a behavioural entity task, and to investigate how to manipulate the initial data for generating new insights and improve the final classification. To the best of our knowledge, this is the first work that combine temporal graph dissection, clustering operation and planning the usage of graph deep learning for detecting malicious entities in cybersecurity domain.

\subsection{Behavioral Node Identification}\label{sec:behavioralnode}

In order to extract and define behavioural entities within the traffic dataset  a pre-processing operation called Behavioral Node Identification is introduced.
In particular, knowing that each record of the dataset represents a communication between two nodes, we map that communication as an edge in the graph. These edges connect an entity source and an entity destination defined respectively by the combination of the source IP (srcip) and source port (srcport), and the destination IP (dstip) and the destination port (dstport) given as features for the considered record.

\begin{figure}[!htbp]
  \centering
    \includegraphics[width=0.6\linewidth]{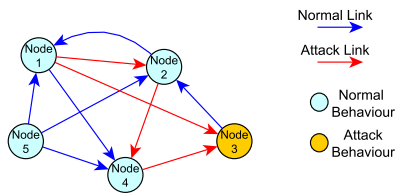}
    \caption{Behavioral Node Identification example}
    \label{fig:behaviour}
\end{figure}

Once obtained the graph, it is important to assign a label to each node in order to define its behavior in the network. In this sense, the information provided by the labelled records of the network traffic dataset are used to label the nodes' behaviour in the network.
The possible nodes' behaviours are two: a normal behaviour identified with the labelled $0$ and an anomalous, or attack, behaviour identified with the label $1$. Each node behaviour is defined by analyzing the nodes' connections, both the incoming and outgoing (Figure \ref{fig:behaviour}). If the malicious connections represent the majority of the node's connections, the node takes the label of anomalous behaviour ($1$), as for example node 3 in the Figure \ref{fig:behaviour}. On the other hand, the node takes the label of normal behaviour ($0$), as for example node 1 in Figure \ref{fig:behaviour}. In case of draw, the node takes the label of normal behaviour ($0$), as for example node $4$.

\subsection{Temporal graph}\label{sec:temporal}
The Node Behavioural Identification can be applied considering the whole network traffic dataset for creating an unique static and monolithic graph, however, with this approach all the dynamics and small interaction between the entities will be overwhelmed by the more frequently and more heavy - in terms of exchanged data - ones. Furthermore, this approach increases the amount of data in the graph requiring more computational resources, making hard the real application of the solution. For example, if a model is trained with a monolithic graph created with $\sim 10$ hours data, for the testing phase, it needs of a comparable structure, i.e. a graph obtained in other $\sim 10$ hours data. This can compromise the usability of the trained model, since in anomaly detection the timing is fundamental in order to mitigate promptly an attack applying countermeasures and avoid the worsening of the situation.

To address these issues, and learn the micro behaviours in the network and detecting the small dynamics within, in this research, the initial dataset is divided in fixed temporal intervals (or temporal snapshot), and in each one of it, it is performed an operation of Node Behaviour Identification, as described in the Section \ref{sec:behavioralnode}. Basically, all the communications in a temporal snapshot are analyzed in order to create a photograph of the network status through a relational graph where the behaviours' entities and their relations are drawn. 

\subsection{Clustering Algorithms}\label{sec:cluster}
In cybersecurity domain, the network traffic datasets are usually characterized by class imbalanced problem. In fact, it is easier to find and generate traffic related with normal activities rather than attack connections. This imbalance problem in the population, between normal and attack entities, can generate loss of classification quality when supervised classification algorithms are used, since the model will be trained with a skewed dataset favoring the classification of a the major-represented class over the under-represented one \cite{guo2008class}. In this scenario, in order to reduce the amount of information in each temporal snapshot, a clustering algorithm is used. The idea is to apply this operation in order to aggregate entities with similar behaviours and homogenize the final population. In particular, we propose to apply this operation just over the normal behavioural entities in order to reduce their population. 

A variety of clustering algorithms can be separated based on the criteria used to create the groups \cite{xu2015comprehensive}, the more common are the hierarchical clustering, the distribution-based clustering, the density-based clustering and the centroid-based clustering. In this work, the analysis is focused over the density-based clustering which represents a group of algorithms where the main parameter to define the clusters is the density difference between zones of the space \cite{kriegel2011density}. This cluster's category uses unsupervised learning methods that identify distinctive groups in the data, based on the idea that a cluster in a data space is a contiguous region of high point density, separated from other such clusters by contiguous regions of low point density\cite{kriegel2011density}. The data points in the separating regions of low point density are typically considered noise/outliers \cite{density_definition}.

In particular, in this analysis, DBSCAN, OPTICS and HDBSCAN techniques are used. The first $2$ algorithms define dense regions using a minimum number of points that must belong to the cluster (\textit{minPts}) and the maximum distance from one point to another for both to be considered neighbours ($\epsilon$). On the other hand, DBSCAN paradigm intends to automatically find a clustering that gives the best stability over $\epsilon$.

After the application of the cluster algorithm, the shape of each temporal graph changes, and so, the corresponding adjacency matrix changes. This aggregation promotes the creation of new clustered entities in which their behaviours are computed by combining (averaging) the single behaviour of the aggregated entities.


\section{Validation}\label{sec:expframwork}

\subsection{Experiment}\label{sec:exp}

In this work, UNSW-NB15 dataset\footnote{https://www.unsw.adfa.edu.au/unsw-canberra-cyber/cybersecurity/ADFA-NB15-Datasets/} is used as input for the \textit{Manipulation phase}. As described in \cite{moustafa2015unsw}, this dataset was created with the aim to improve the existing benchmark datasets, which are not able to describe the comprehensive representation of network traffic and attack scenarios.
The UNSW-NB15 dataset contains real normal and synthetic abnormal network traffic generated in the synthetic environment of the University of New South Wales (UNSW) cyber security lab.
The UNSW-NB15 dataset is characterized by $9$ attack families and normal traffic generated in two distinct capture-day. In particular, in this study, the connections are labelled as $0$ if they are normal traffic or $1$ if they are attack traffic.

The \textit{Manipulation phase} starts fixing the value for the temporal dissection operation, in particular, for this work, $600$ seconds ($10$ minutes) is chosen. Then, from each of the extracted temporal snapshots a behavioural graph is created in order to detect the entities and their relations within the network. For each node, a feature vector that describes its behaviour, is extracted and it is used as input for the clustering operation. This clustering process allows us to reduce the amount of data and to mitigate the imbalance problem as presented in Section \ref{sec:temporal}. In particular, $3$ clustering algorithms - OPTICS, DBSCAN and HBDSCAN - are applied separately, studying how their parameters affect their effectiveness. In particular, the OPTICS and the DBSCAN algorithm are implemented both with \textit{minPts} value of $2$ and with $3$ different $\epsilon$ values: $0.2$, $0.5$ and $0.8$. The HDBSCAN instead is applied again with \textit{minPts} equals $2$ and with a minimum size of clusters of $5$, and a euclidean metric to compute the distance between the instances is used.

\subsection{Results}\label{sec:results}

\begin{figure}[!htbp]
  \centering
    \includegraphics[width=0.85\linewidth]{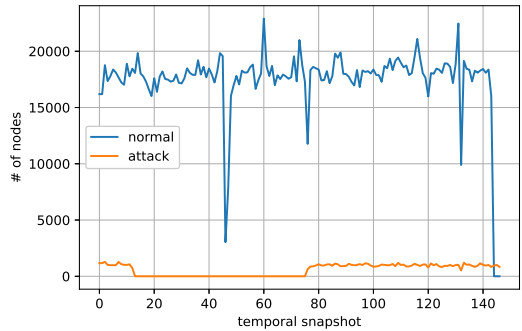}
    \caption{Initial population in each temporal size}
    \label{fig:temporalpopulation}
\end{figure}

Choosing fixed temporal intervals of $600$ seconds, $147$ temporal snapshots are generated, as shown in Figure \ref{fig:temporalpopulation}. Each temporal snapshot is characterized by a very large number of nodes, where the amount of normal entities overwhelm the amount of anomalous one. In fact, in all the extracted $147$ temporal snapshots, the average number of normal nodes in a snapshot is more than $17,000$, for an overall amount of $2,583,605$, while the average attack population is about $545$, for a global value of $80,174$. Furthermore, in several temporal snapshots there are not attack nodes, as for example between the $17$th and the $77$th snapshots.

Table \ref{tab:clusteredpopulation} shows the overall clustered population generated by each algorithm. This overall value is computed by summing the number of normal nodes in all temporal snapshots after the execution of the cluster algorithms, where the low point density regions are discarded, as they are considered as noise/outliers.

\begin{figure}[!htbp]
  \centering
    \includegraphics[width=0.85\linewidth]{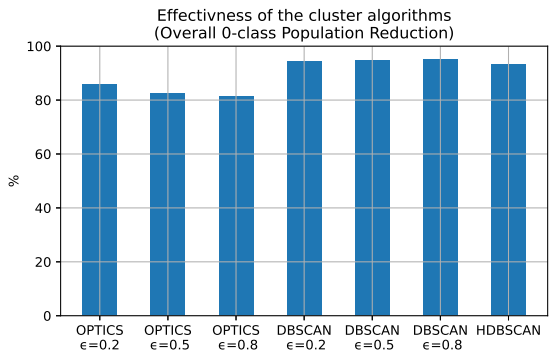}
    \caption{Clustered Population}
    \label{fig:clusteredpopulation}
\end{figure}

It should be noted that from an initial overall amount of $2,583,605$ normal nodes, the cluster algorithms reduce them in a range between $479,938$ and $127,346$, i.e. reducing the class of about $81$\% - $95$\% (Figure \ref{fig:clusteredpopulation}). Nevertheless, for the OPTICS datasets, the new normal population still represent more than $80$\% of the whole population.
These cluster methods are not applied over the attack class, as explained in Section \ref{sec:temporal}, so the final attack population is invariant ($80,174$ nodes)

Table \ref{tab:clusteredpopulation} shows also that a fixed value of $\epsilon$ parameter generates different results according to the considered clustering algorithm. In particular, increasing the $\epsilon$ value increases the ability of the OPTICS algorithm to aggregate the samples in dense regions, reducing the number of noise/outliers. The same highest value of $\epsilon$ produces a reverse effect in the DBSCAN, in fact, it decreases its ability to create dense regions, increasing the number of noise/outliers and reducing the final population.
 
\begin{table}[]
\def\arraystretch{1.5}
\centering
\begin{tabular}{cccc}
\hline
\textbf{Method} & \textbf{$\epsilon$} & \textbf{\begin{tabular}[c]{@{}c@{}} normal entities\\after clustering\end{tabular}} & \textbf{\begin{tabular}[c]{@{}c@{}} clustered normal entities\\over the whole dataset\end{tabular}} \\ \hline
\multirow{3}{*}{OPTICS} & 0.2 & 369,931 &82.19 \%   \\
 & 0.5 & 447,026 &84.79 \%  \\
 & 0.8 & 479,938 &85.69 \% \\\hline
\multirow{3}{*}{DBSCAN} & 0.2 & 141,935 &63.90 \% \\
 & 0.5 & 137,903 &63.24 \% \\
 & 0.8 & 127,346 &61.37 \%  \\\hline
HDBSCAN & - & 176,645 &68.78 \%  \\ \hline
\end{tabular}
\caption{Clustering effects}
\label{tab:clusteredpopulation}
\end{table}

\subsection{Learning guidelines}\label{sec:gcn}

Once the \textit{Manipulation phase} is ended, the information is ready for training learning models and achieve the attack behavioural classification task related with the \textit{Learning phase}. In particular, in order to fully exploit the extracted graph relations highlighted in the first phase, it is planned to use the Graph Convolutional Network. These models, introduced in \cite{kipf2017semi}, are able to learn the local and global structural patterns of a graph through a convolution operation as well as happens in the Convolution Neural Network (CNN).

The convolution operation can be defined through the Equation \ref{eq:conv}, where $x \in \mathbb{R}^n$ represents a scalar vector (the scalar vector for every node in the graph), $U$ is the matrix of eigenvectors of the Laplacian of the graph and $g_{\theta}(\Lambda)$ represents the filter in the Fourier domain. Nevertheless, solving this equation can be computationally complex and unreachable, even more prohibitively expensive for large graphs.

\begin{equation}\label{eq:conv}
y = g_\theta*x = Ug_{\theta}(\Lambda)U^{T}x
\end{equation}

In this sense, in \cite{defferrard2016convolutional} a promising solution to the Equation \ref{eq:conv} is obtained parametrizing the term $g_{\theta}(\Lambda)$ as a polynomial function that can be computed recursively. In particular, Chebyshev polynomials with $k$ degree are used. Kipf et al. \cite{kipf2017semi} demonstrated that a good approximation can be reached by truncating the Chebichev polynomial to get a linear polynomial ($k$ = 1) and by performing a renormalization trick in order to avoid numerical instabilities and vanishing gradients.

For the validation of the \textit{Learning phase}, the idea is to exploit the highlighted graph-based relations applying $2$ GCNs models, the simple one introduced in \cite{kipf2017semi}, and its higher version based on a different approximation of the Chebychev polynomials ($k > 1$), in order to highlight how the relational information affect the final classification.

\section{Conclusions}\label{sec:conclusions}

In this research, a two phased approach for converting a network traffic analysis into a behavioural entity classification task is presented. In particular, this work introduces and validates the first phase of this approach called \textit{Manipulation phase}, and draws several guidelines for the next phase, called \textit{Learning phase}

The main idea of the first phase is to understand and manipulate correctly the input data regardless the chosen classification model. In particular, a method for converting the network traffic time series into graph-based structures is presented, as well as a temporal dissection operation used to uncover micro-dynamics that in a complex system can be overwhelmed by the most recurrent ones. Furthermore, $3$ clustering techniques - OPTICS, DBSCAN and HDBSCAN - for reducing noisy data and aggregate similar behaviour, are tested and compared using a relevant cybersecurity dataset (UNSW15). A preliminary study shows how their effectiveness changes according to several parameters, as well as the DBSCAN is more proned to create regions with lower density that are considered as noise/outliers. 
Finally, $3$ GCNs' architectures are presented as relevant candidate for implementing the \textit{Learning phase}

\section*{acknowledgement}

This work has been partially supported by the Basque Country Government under the ELKARTEK program, project TRUSTIND (KK-2020/00054).

\bibliographystyle{IEEEtran}
\bibliography{jnic2021}
\end{document}